\documentclass[runningheads]{llncs}
\usepackage[square, numbers]{natbib}

\usepackage{graphicx}
\usepackage{booktabs}
\usepackage{comment}
\usepackage{xcolor}
\usepackage{amsmath}
\usepackage{amssymb}
\usepackage{makecell} 
\usepackage{hyperref}
\usepackage{subcaption}
\usepackage[T1]{fontenc}

\begin{document}

\title{HyperMM : Robust Multimodal Learning with Varying-sized Inputs}


\author{Hava Chaptoukaev\inst{1} \and Vincenzo Marcian\'o\inst{1,2} \and Francesco Galati\inst{1} \and Maria A. Zuluaga\inst{1}}
\authorrunning{H. Chaptoukaev et al.}
\institute{EURECOM, Data Science Department, France 
\and School of Biomedical Engineering \& Imaging Sciences, King’s College London, UK
\\
\email{\{chaptouk,marciano,galati,zuluaga\}@eurecom.fr}}

\maketitle              

\begin{abstract}
Combining multiple 
modalities carrying complementary information through multimodal learning (MML) has shown considerable benefits for diagnosing multiple pathologies. However, the robustness of multimodal models to missing modalities is often overlooked. Most works assume modality completeness in the input data, while in clinical practice, it is common to have incomplete modalities. Existing solutions that address this issue rely on modality imputation strategies before using supervised learning models. These strategies, however, are complex, computationally costly and can strongly impact subsequent prediction models. Hence, they should be used with parsimony in sensitive applications such as healthcare. We propose HyperMM, an end-to-end framework designed for learning with varying-sized inputs. Specifically, we focus on the task of supervised MML with missing imaging modalities without using imputation before training. We introduce a novel strategy for training a \textit{universal} feature extractor using a conditional hypernetwork, and propose a permutation-invariant neural network that can handle inputs of varying dimensions to process the extracted features, in a two-phase \textit{task-agnostic} framework. We experimentally demonstrate the advantages of our method in two tasks: Alzheimer's disease detection and breast cancer classification. We demonstrate that our strategy is robust to high rates of missing data and that its flexibility allows it to handle varying-sized datasets beyond the scenario of missing modalities. We make all our code and experiments available at \href{https://github.com/robustml-eurecom/hyperMM}{github.com/robustml-eurecom/hyperMM}.

\keywords{Multimodal learning \and Missing modalities \and Multi-resolution data}
\end{abstract}

\section{Introduction}
\label{Intro}
Multimodal imaging techniques are widely used both in clinical practice and medical research. Simultaneous acquisition and analysis of multiple imaging modalities, such as Emission Tomography (PET), Computed Tomography (CT), or Magnetic Resonance Imaging (MRI), has shown to be beneficial in the diagnosis of Alzheimer's disease~\cite{teipel2015multimodal}, or detection of cancers~\cite{tempany2015multimodal}, among others. Accordingly, deep learning methods designed to learn from multimodal medical images~\cite{fu2021multimodal, odusami2023machine}, {and more generally multimodal medical data~\cite{sun2023scoping}}, have seen rapid growth. This development has been favored by the emergence of multimodal learning (MML), a field of machine learning combining modalities from various sources that depict a single subject from multiple views, thus providing both shared and complementary information. MML has shown considerable advantages in multiple domains~\cite{baltruvsaitis2018multimodal, xu2023multimodal}. However, most current models~\cite{wu2018multimodal, huang2019diagnosis, zhang2020multi, venugopalan2021multimodal, zuo2021multimodal} assume completeness of the training and testing data, which is rare for real-world datasets. In particular, in routine clinical practice obtaining several modalities for the same subject is not a standard. Incomplete datasets can occur for multitudes of reasons including databases fusion, unavailability of acquisition material, or simply patient refusal to partake in specific examinations. As a result, having varying numbers of modalities per patient is common, which results in multimodal datasets where one or more modalities can be missing. This makes MML challenging as it prevents the straightforward use of the existing methods. Moreso, multimodal models trained on complete datasets become unusable (without complex additional processing steps) if modalities are missing at testing time, which severely restricts their usage to complete samples only. Therefore, the robustness of multimodal models to missing modalities is of paramount importance for the use of MML in real-life applications.

\subsection{Related work} MML aims to build models that process and combine information from multiple sources~\cite{baltruvsaitis2018multimodal}, i.e. multiple modalities. The most prominent way to combine multi-source information resides in fusion methods that can be classified in three categories: early fusion, mid-level fusion, and decision-level fusion of modalities~\cite{xu2023multimodal}. In practice, summation and averaging are common and straightforward techniques used for fusion. However, when modalities are missing, these operations are impossible for early and mid-level fusion in classical multimodal architectures. They are usually not designed to handle varying-sized inputs and fail to account for missing data. 

A vast majority of existing solutions to missing modalities in supervised learning consists of first training a generative model on a complete dataset, and using it to impute missing modalities before learning a discriminative model for prediction~\cite{cai2018deep, kim2020multi, sun2021semi, zhang2023unified}. This approach has considerable limitations in practice. First, an unreasonable number of samples may be needed for training a good missing-modality imputation model. For instance, generative adversarial networks (GANs)~\cite{isola2017image, zhu2017unpaired}, often used for image generation and imputation, can typically require up to $10^6$ samples for efficient training~\cite{karras2020training}. This considerably limits their uses in medical applications where data is often scarce. Second, the complexity of the prediction model strongly depends on the choice of the imputation model. The imputer and predictor networks need to be adapted to each other~\cite{le2021sa, lu2024theory}, which can be difficult to ensure in practice. Some studies~\cite{suo2019metric, wang2023multi} address this limitation by focusing on jointly learning the modality imputation and prediction tasks, but these models rely on complex and computationally costly training strategies. Lastly, poorly imputed data can compromise the interpretability 
of subsequent predictors~\cite{shadbahr2023impact}, which is a crucial aspect to consider in sensitive applications, such as healthcare, where it can lead to incorrect conclusions about the impact of a feature on the outcome. 
 
Some recent works have proposed handling missing data without using imputation~\cite{parthasarathy2020training, zhou2023incomplete, chen2024unified}. Instead of directly imputing the missing modalities, they replace them with \textit{dummy} inputs,  such as a constant or generated data (e.g., zeros or Gaussian noise), and then learn to ignore these during training. In contrast, we propose to learn with varying-sized inputs to avoid model degradation caused by poor imputations or the presence of \textit{dummy} data.

\subsection{Contributions} In this work, we address the limitations of existing methods by proposing an end-to-end \textit{imputation-free} strategy for multimodal supervised learning with missing imaging modalities. Building on conditional hypernetworks~\cite{ha2016hypernetworks}, we formulate a novel strategy for training a \textit{universal} modality-agnostic feature extractor using a large pre-trained network.  We then reformulate the problem of predicting multimodal observations with missing modalities as one of predicting \textit{sets} of observations of varying size, thus relaxing the requirement of fixed-dimensional data inputs of most machine learning models. We implement this approach through a permutation-invariant neural network~\cite{zaheer2017deep}, allowing the mid-level fusion of varying-sized multimodal inputs, hence eliminating the need to impute~\cite{cai2018deep, sun2021semi, zhang2023unified} or mask missing modalities using dummy data~\cite{parthasarathy2020training, zhou2023incomplete, chen2024unified} as done in previous works. By combining these elements into a two-step training framework, we formulate HyperMM, a novel \textit{task} and \textit{model agnostic} strategy for MML from incomplete datasets, without the use of imputation or dummy data in the training process. To the best of our knowledge, our work is the first proposing such an approach for multimodal learning with missing modalities.

\section{Methodology}
\subsection{Overview of the method}
\label{sec:overview}
We consider a dataset $\mathcal{D}$ of $n \in \mathbb{N}$ independent input and output pairs such that $\mathcal{D}:=\{(X_1,Y_2), \dots, (X_n, Y_n)\}$, and for which the goal is to predict $Y$ given $X$. Each $X := \{x_1, \dots, x_d\}$ corresponds to a $d$-modal observation, where each $x_i$ represents one of the available modalities. Let us now introduce the indicator vector $v \in \{0,1\}^d$ to denote the positions of missing modalities in $X$, such that $v_i=1$ if $x_i$ is missing, and $0$ otherwise. The observed data of $X$ can be expressed as $X_{obs} = (1 - v) \odot X + v \odot \texttt{na}$, where $\odot$ is the term-by-term product. In this setting, the learning goal becomes the prediction of $Y$ given $X_{obs}$. 

\begin{figure}[t]
\includegraphics[width=\textwidth]{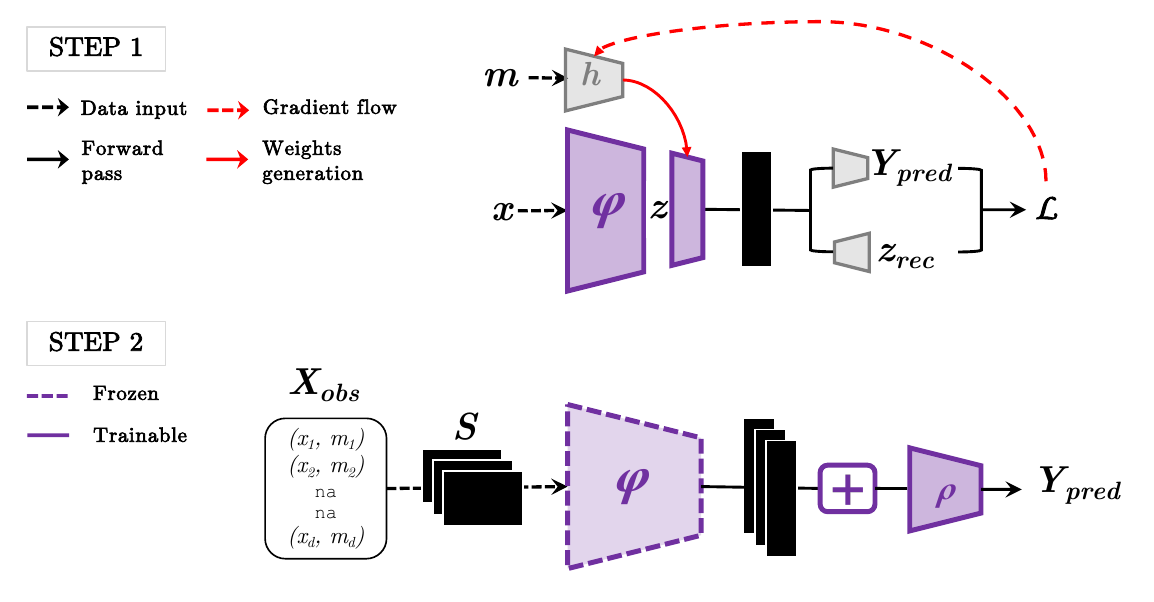}
\caption{Overview of our HyperMM framework. A network $\varphi$ is trained to extract features from any modality in $\mathcal{D}$ by jointly optimizing feature reconstruction and unimodal prediction (step 1). The learned $\varphi$ is frozen, and used to process multimodal inputs, the latent features are then aggregated and processed through a network $\rho$ for prediction (step 2).}
\label{fig:model}
\end{figure}

We intend on learning without the use of any form of imputation of missing modalities, and therefore, with entries of different dimensions. However, standard machine learning models, including MML models, are built to handle data inputs of a fixed size. In contrast, we aim to learn a sum-decomposable function $f$ of the form $f = \rho(\sum \varphi(x_i))$, operating on \textit{sets} and thus relaxing the requirement of fixed-dimensional data. We propose a two-step framework that we call HyperMM to implement our method. Figure~\ref{fig:model} presents an overview of our strategy. In a first step, we learn a neural network $\varphi$ that can extract features from any modality present in $\mathcal{D}$. Then in a second step, we freeze the learned $\varphi$, use it to encode each element of $X_{obs}$, and feed the combination of the encoded inputs to a classifier $\rho$ through a permutation-invariant architecture. 

\subsection{Universal Feature Extractor}
A single network $\varphi$ that can encode all observed modalities in $\mathcal{D}$ is a requirement for learning a set function as described in Sec.~\ref{sec:overview}. We propose to achieve this by first learning such a universal feature extractor $\varphi$ using a conditional hypernetwork~\cite{ha2016hypernetworks}. In this first step, we train a network on all available images $x$ in the dataset, without any modality pairing. As illustrated in Figure~\ref{fig:model}, we introduce an \textit{auxiliary} network $h$ that takes as input $m$, the modality identifier corresponding to the image $x$, and generates conditional weights for the last layer of the encoder $\varphi$. By doing so, the last feature extraction step is different for each modality but still performed by the same network. Specifically, modality-specific layers are generated through a common hypernetwork, which facilitates information sharing across modality-specific layers. 

In practice, our universal feature extractor $\varphi$ can be implemented using transfer learning and networks pre-trained on natural images such as VGGs~\cite{simonyan2014very}. First, we use the pre-trained encoding layers of a VGG to extract features from our dataset. Then, we adapt the obtained general features into medical ones by training an additional layer on top of the VGG extractor, that is conditioned using the auxiliary network $h$. By stacking these elements together, we obtain our universal feature extractor $\varphi$ that is adapted to the modalities of our dataset.

To ensure that the features learned by $\varphi$ are relevant, the network is trained to both predict $y$ from the single modality images, and reconstruct $z$, the features outputed by the second-to-last layer of $\varphi$. This is achieved by optimising a loss function of the form $\mathcal{L} = \mathcal{L}_{MSE} + \mathcal{L}_{CE}$, where $\mathcal{L}_{MSE}$ denotes the mean squared error between $z$ and $z_{rec}$, and $\mathcal{L}_{CE}$ the cross-entropy loss between $y$ and $y_{pred}$. This optimisation loss has been chosen by cross-validation, as it yielded better performances than optimising on the classification or reconstruction only.

\subsection{Permutation Invariant Architecture}
Once we have learned $\varphi$, we freeze it, and use it to implement a permutation invariant network for supervised MML with missing modalities. To do so, we define $S$, the set representation of the $q = |S|$ observed elements of $X_{obs}$, such that $S := \{s_1, \dots, s_q\}$, with $q \leq d$. Each element $s_j$ is represented as a tuple $(x_i, m_i)$ consisting of an observed modality $x_i$, and the corresponding modality identifier $m_i$. This reformulation allows observations of varying dimensions. Thereby, it does not require nor expects all observations to have the same number of elements and it fully allows observations with missing modalities. A $d$-modal observation $X_{obs}$ containing \texttt{na} values can simply be expressed as a set $S$ of size $q \leq d$ where the \texttt{na} values are not represented anymore. 

Using this definition, we leverage on the findings of \cite{zaheer2017deep}, who proposed a learning framework that considers permutation invariant functions operating over sets. We reformulate our learning goal as one of learning a set function $f$ of the form
\begin{equation}
\label{eq1}
f(X_{obs}) = \rho \left(\sum_{s_k \in S} \varphi(s_k) \right),
\end{equation}
where the function $\varphi : \mathbb{R} \times \{r\times r\}^d \to \mathbb{R}^{d_l}$ corresponds the encoder obtained from the pre-training phase, the function $\rho: \mathbb{R}^{d_l} \to \mathbb{R}$ is implemented as neural network, $r$ is the size of each image and $d_l \in \mathbb{N}^+$ is the dimensionality of the latent space of $\varphi$. 

As illustrated in Figure~\ref{fig:model}, a given observation $X_{obs}$ with missing modalities is encoded as a set $S$. Each element $s_k \in S$ is then transformed into a representation $\varphi(s_k):=\varphi(x_i|m_i)$ through the frozen network $\varphi$ conditioned by the modality identifier $m$. The representations $\varphi(s_k)$ are aggregated using a permutation invariant operation such as the sum, the mean or the maximum. The aggregation is processed through the network $\rho$, which allows to predict the target $Y$ corresponding to the input $X_{obs}$. The proposed architecture interprets each observation $S$ of a dataset as a set of unordered modalities, where all information available in $X_{obs}$ is conserved and no new information, such as imputed images, is added. By transforming individual elements $s_k$ of $S$ at a time and then aggregating the transformations, our network encodes sets of arbitrary sizes into a fixed representation $\sum \varphi(s_k)$. This aspect is particularly relevant and further justifies handling our dataset with missing modalities as unordered sets.

Our permutation invariant model is learned by optimising the loss function
\begin{equation}
\label{eq2}
    \mathcal{L}(\theta) := \mathbb{E}_{(S, Y) \in \mathcal{D}}\Bigl[ \ell \Bigl(Y, \rho_{\theta}\Bigl( \sum_{s_k \in S}\varphi(s_k)\Bigr)\Bigr)\Bigr],
\end{equation}
where $\rho$ is parametrised by $\theta$, and $\ell$ is the cross-entropy loss. As $\varphi$ is optimised in the pre-training step, its weights are not updated in this step.

\section{Experiments}
\subsection{Alzheimer's Disease Detection}
\label{sec:AD}
In a first application, we illustrate the performances of HyperMM and its robustness to missing modalities on the task of binary classification of Alzheimer's disease (AD) using multimodal images from the ADNI dataset~\cite{mueller2005ways}. We select a subset of $300$ patients for which both T1-weighted MRIs and FDG-PET images are available, resulting in $165$ cognitively normal (CN) and $135$ AD observations. Before learning, all the samples are skull stripped using HD-BET~\cite{isensee2019automated}, resampled through bicubic interpolation to set an uniform voxel size, standardised, and normalised using min-max scaling. \\

\noindent\textbf{Baselines.} We first evaluate the advantages of our strategy for MML with complete data. We compare the performances of HyperMM against: \textbf{Uni-CNN}, an unimodal CNN as implemented by~\cite{liang2021alzheimer}; \textbf{Multi-CNN}, a multimodal CNN as proposed by~\cite{venugopalan2021multimodal}; and \textbf{Multi-VAE}, a multimodal VAE~\cite{wu2018multimodal} that we adapt for classification. Then, we compare our method against state-of-the-art techniques for MML with missing modalities in two scenarios: complete MRIs $+50\%$ of PETs available for training and testing, and complete PETs $+50\%$ of MRIs available. Specifically we compare to: \textbf{pix2pix}, a strategy where an image-to-image translation model~\cite{isola2017image} is trained on the subset of the training data containing only modality-complete samples, is then used to impute the missing modality of the incomplete data, and once imputed the data is classified using a Multi-CNN; and \textbf{cycleGAN}, the same strategy, only using a cycleGAN~\cite{zhu2017unpaired} for imputation. \\

\begin{figure}[t]
\includegraphics[width=\textwidth]{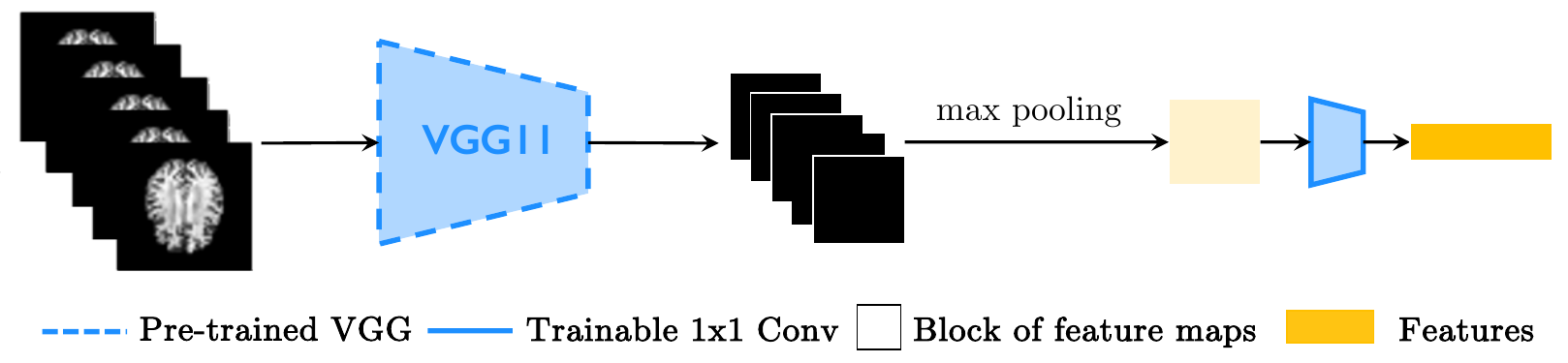}
\caption{Feature extraction strategy used in the ADNI baselines (see~\cite{liang2021alzheimer}). All 2D slices of one 3D volume are fed to a VGG11. A 1D max pooling on the slice dimension is applied to the resulting feature blocks to obtain a single block per 3D image. The latter is passed through a $1 \times 1$ convolution layer to obtain AD-specific features that can then be fed to a classifier.}
\label{fig:fe}
\end{figure}

\noindent\textbf{Implementation details.} We randomly split the data into train, validation and test sets with a 6:1:3 ratio on the patient-level, and repeat all experiments 3 times.
For simplicity and fairness, we use the same feature extraction strategy (Figure~\ref{fig:fe}) in all baselines, following~\cite{liang2021alzheimer}.
Specifically, 3D MRI and PET images are processed as batches of 2D slices that are each fed to a pre-trained frozen VGG11~\cite{simonyan2014very} feature extractor. We feed all 2D slices of a 3D volume to the VGG, and apply a 1D max pooling on the slice dimension to the resulting feature blocks to obtain a single block per 3D image. The resulting block is passed through a $1 \times 1$ convolution layer after the pre-trained VGG encoder, allowing us to adapt the pre-trained features into AD-specific ones. 
This corresponds to the training of the $\varphi$ network in the step 1 of our framework, where we simply make the last $1 \times 1$ convolution layer conditional. In step 2, the $\rho$ network is implemented by 3 linears layers separated by ReLU activations. All models are implemented with PyTorch, and trained on an Nvidia TITAN Xp GPU for a maximum of 100 epochs using an early stopping strategy, where training stops after 10 iterations without a decrease in the validation loss. We use a batch size of 1 and an Adam optimiser with an initial learning rate of $1\mathrm{e}{-4}$.
\\

\begin{table}[t]
\caption{Performances (mean$\pm$std) on the ADNI dataset. \textbf{Bold} means best.}\label{tab1}
    \begin{center}
        \resizebox{\textwidth}{!}{
        \begin{tabular}{ccccccc}
        \hline  
         &  Acc. ($\uparrow$) & AUC ($\uparrow$) & F1 ($\uparrow$) & Prec. ($\uparrow$) & Rec. ($\uparrow$) & Time ($\downarrow$)\\
        \hline
        \textbf{Complete unimodal} \\ 
        Uni-CNN PET  & 0.61$\pm$.05 & 0.58 $\pm$.05 & 0.58$\pm$.06 & 0.65$\pm$.06 & 0.31$\pm$.05 & $<$ 20 min \\ 
        Uni-CNN MRI  & 0.71$\pm$.02 & 0.69$\pm$.02 & 0.58$\pm$.02 & \textbf{0.85$\pm$.03 }& 0.43$\pm$.05 & $<$ 20 min \\ 
        \hline
        \textbf{Complete multimodal} \\ 
        Multi-VAE classifier & 0.66$\pm$.03 & 0.65$\pm$.03 & 0.54$\pm$.04 & 0.74$\pm$.04 & 0.41 $\pm$.03 & $<$ 30 min \\
        Multi-CNN & 0.70$\pm$.02 & 0.70$\pm$.01 & 0.67$\pm$.01 & 0.67$\pm$.02 & 0.68$\pm$.02 & $<$ 30 min \\ 
        HyperMM w/o 2-steps (ours) & 0.62$\pm$.03 & 0.61$\pm$.02 & 0.53$\pm$.02 & 0.61$\pm$.03 & 0.46$\pm$.03 & $<$ 20 min \\ 
        HyperMM w/ 2-steps (ours) & \textbf{0.74$\pm$.02} & \textbf{0.73$\pm$.02} & \textbf{0.70$\pm$.01} & 0.70$\pm$.02 & \textbf{0.70$\pm$.02} & $<$ 1 h\\ 
        \hline
        \textbf{100\% MRI + 50\% PET} \\ 
        pix2pix  & 0.65$\pm$.02 & 0.64$\pm$.02 & \textbf{0.62$\pm$.02} & \textbf{0.62$\pm$.03} & \textbf{0.61$\pm$.02} & $>$ 14+1 h \\ 
        cycleGAN & 0.62$\pm$.09 & 0.60$\pm$.07 & 0.57$\pm$.07 & 0.61$\pm$.08 & 0.54$\pm$.08 & $>$ 30+1 h\\
        HyperMM (ours) & \textbf{0.67$\pm$.02}& \textbf{0.66$\pm$.02} & 0.61$\pm$.03 & 0.61$\pm$.03 & 0.61$\pm$.03 & $<$ 1 h\\
        \hline 
        \textbf{100\% PET + 50\% MRI} \\ 
        pix2pix & 0.62$\pm$.04 & 0.62$\pm$.03 & 0.53$\pm$.03 & 0.61$\pm$.05 & 0.48$\pm$.05 & $>$ 14+1 h \\
        cycleGAN & 0.62$\pm$.09 & 0.59$\pm$.1 & 0.47$\pm$.07 & 0.60$\pm$.07 & 0.39$\pm$.07 & $>$ 30+1 h \\
        HyperMM (ours) & \textbf{0.64$\pm$.02} & \textbf{0.63$\pm$.02} & \textbf{0.61$\pm$.02} & \textbf{0.61$\pm$.03} & \textbf{0.61$\pm$.03} & $<$ 1 h\\ 
        \hline
        \end{tabular}
        }
    \end{center}
\end{table}
\begin{figure}[t]
\includegraphics[width=\textwidth]{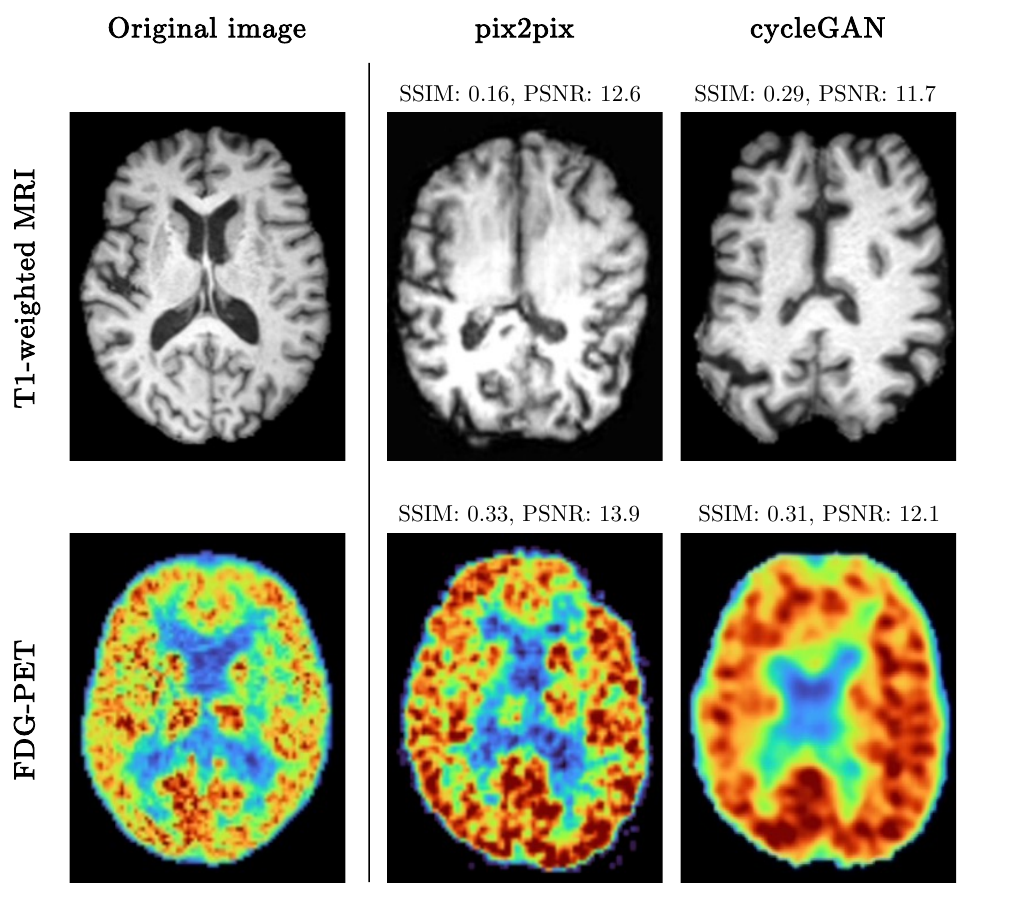}
\caption{Examples of real and imputed slices of MRI and PET images for one patient. While the PET reconstructions (bottom right) translated from the corresponding MRI (top left) are reasonably similar to the original PET image (bottom left), the MRI reconstructions (top right) translated from the low-resolution PET (bottom left) are much less consistent with reality (top left).}
\label{fig:ex}
\end{figure}
\noindent\textbf{Results.} Performances of all models are reported in Table~\ref{tab1}. Several observations can be drawn from these results. First, MML shows considerable improvements over unimodal baselines. In particular, HyperMM achieves the best performances for binary classification of AD using complete multimodal data and considerably improves the F1-score, recall metric, and precision/recall balance. Second, MML with missing modalities still achieves better results than unimodal models. Notably, HyperMM trained on MRIs available even for only 50\% of the patients performs better than an unimodal model trained on PETs only. Inversely, having access to PETs for 50\% of the patients improves the F1-score and recall of learning from MRIs only. Third, HyperMM outperforms state-of-the-art strategies on MML with missing modalities. While GAN-based strategies can handle missing PETs in the input data, they are considerably less efficient in terms of precision/recall balance when the missing modality is MRI. In this scenario, the missing high-resolution MRIs need to be translated from the available low-resolution PETs before learning. This limitation is further illustrated in Figure~\ref{fig:ex}. While PET imputation yields realistic images, the imputed MRIs are of poor quality: they suffer from important structural deformations and a great loss of information (as highlighted by the SSIM and PSNR scores between the imputations and the original images). In contrast, as HyperMM does not rely on any imputation, it performs well in both scenarios, and trains in significantly less time than competitors. Lastly, these results highlight the importance of the pre-training and conditioning step of the HyperMM framework. \\

\noindent\textbf{Discussion.} The results illustrate how HyperMM tackles the main limitations of existing methods. First, as our model does not require training an imputation model prior to prediction, it does not call for the large amounts of data typically required for training GANs efficiently. The results observed in Table~\ref{tab1} highlight the poor performances of cycleGAN for translating PETs into MRIs, which could be due to insufficient training data. Second, our model is agnostic to the missing modality, whereas the prediction and imputation quality in other approaches strongly depends on it, as highlighted by our experiments. Indeed, because HyperMM bypasses the imputation step altogether, our approaches eliminates the need to ensure that the imputer and predictor are adapted to each other. This, in turn, leads to drastically reduced computing time and learning complexity. Lastly, as our method does not employ any imputed or dummy data, it avoids model degradation caused by poor imputations or noisy data.

\subsection{Breast Cancer Classification}
In a second application, we demonstrate the flexibility of HyperMM and its benefits for learning with varying-sized datasets, beyond the scenario of missing modalities. We investigate the usage of HyperMM for the slightly different task of analysing multi-resolution histopathological images. Because potential tumors are typically acquired at multiple magnification levels, the numbers of samples per patients in histopathology datasets are often highly varying. We perform binary classification of breast cancer using histopathological images from the BreaKHis dataset~\cite{spanhol2015dataset}.  BreaKHis contains multiple images per sample (i.e. patient) of benign or malignant tumors observed through different microscopic magnifications: 40$\times$, 110$\times$, 200$\times$, and 400$\times$. We select a balanced subset of the data composed of samples of 24 benign and 29 malignant tumors, resulting in 5,575 images in total. We use the images as they are for learning, and do not perform any pre-processing or data augmentation. 

In clinical practice, pathologists combine the complimentary information present in images captured under different magnifications in order to make a patient-level decision. Nonetheless, most current learning approaches consist of magnification-specific models, due to the difficulty of processing images of different natures with a single model. Moreover, because the number of available images can vary a lot from one patient to another, traditional algorithms cannot be applied at the patient-level. Existing methods rather predict from individual images, and later combine the predictions in order to form a global decision. Instead, we propose to tackle this problem using HyperMM, conditioning the universal feature extractor on the different magnification levels. We classify tumors at patient-level by combining all available images during training directly. \\

\begin{figure}[t]
\includegraphics[width=\textwidth]{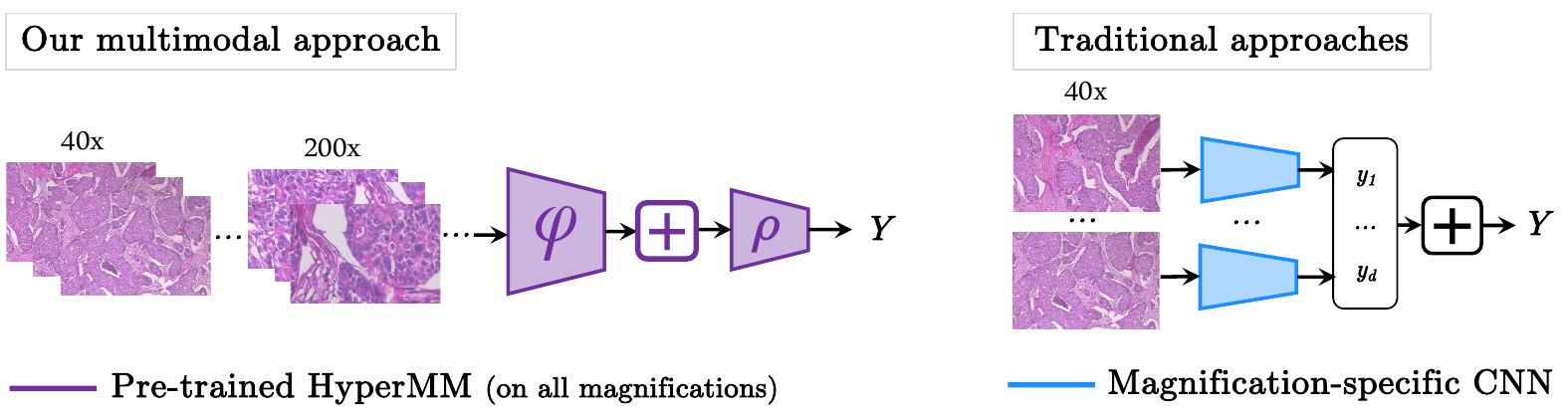}
\caption{Comparison of decision strategies for patient-level tumor classification. Our method (left) enables the combination of a subject's available images during training, regardless of the magnification level to obtain a patient-level decision. In opposition, traditionnal approaches (right) make prediction on the image-level, and combine the final predictions to obtain a patient-level decision.}
\label{fig:breakhis}
\end{figure}

\noindent\textbf{Baselines.} We evaluate the benefits of HyperMM for learning from histopathology data, and compare its performances with: \textbf{CNN}, where a magnification-specific CNN is trained to classify tumor types from individual images, and patient-level prediction is obtained by averaging the classification scores of individual images~\cite{spanhol2015dataset}; and \textbf{Incremental-CNN}, in which a magnification-agnostic CNN is trained by incrementally updating its weights on successive batches of 40$\times$, 100$\times$, 200$\times$ then 400$\times$ magnifications, as proposed in~\cite{mayouf2022curriculum}. The patient-level decision is obtained similarly to the previous baseline. The differences between our approach and traditional ones are further illustrated in Figure~\ref{fig:breakhis}.\\

\noindent\textbf{Implementation details.} We randomly split the data into train and test sets with a 8:2 ratio at the patient-level, and repeated all experiments 5 times. We use a pre-trained VGG11~\cite{simonyan2014very} feature extractor for all baselines, and adapted the features to our application by adding a $1 \times 1$ convolution block on top of the frozen VGG encoder. All models are trained for a maximum of 50 epochs using an early stopping strategy such that training stops after 10 iterations without a decrease in the validation loss. We train the model with an Adam optimiser with an initial learning rate of $1\mathrm{e}{-4}$. We use a batch size of 16 for image-level baselines (i.e. CNN and Incremental-CNN) and 1 for HyperMM.
\\

\begin{table}[t]
\caption{Performances (mean $\pm$ std) on the BreaKHis dataset. \textbf{Bold} means best.}
\label{tab2}
    \begin{center}
        \resizebox{\textwidth}{!}{
        \setlength{\tabcolsep}{2pt}
        \begin{tabular}{ccccccc}
        \hline
         &  Acc. ($\uparrow$) & AUC ($\uparrow$) & F1 ($\uparrow$) & Prec. ($\uparrow$) & Rec. ($\uparrow$) \\
        \hline
        \textbf{Magnification-specific} \\ 
        CNN 40$\times$ & 0.83$\pm$0.07 & 0.81$\pm$0.07 & 0.83$\pm$0.06 & 0.85$\pm$0.08 & 0.83$\pm$0.08\\ 
        CNN 100$\times$ & 0.85$\pm$0.08 & 0.85$\pm$0.08 & 0.87$\pm$0.06 & 0.85$\pm$0.07 & 0.90$\pm$0.07 \\ 
        CNN 200$\times$ & 0.84$\pm$0.07 & 0.84$\pm$0.09 & 0.84$\pm$0.05 & 0.80$\pm$ 0.11 & 0.90$\pm$ 0.09 \\ 
        CNN 400$\times$ & 0.83$\pm$0.09 & 0.83$\pm$0.09 & 0.85$\pm$0.10 & 0.88$\pm$0.11 & 0.83$\pm$0.15 \\ 
        \hline
        \textbf{Magnification-agnostic} \\ 
        Incremental-CNN & 0.89$\pm$0.11 & 0.88$\pm$0.12& 0.90$\pm$0.10& 0.88$\pm$0.12 & \textbf{0.93$\pm$0.09}\\
        HyperMM (ours) & \textbf{0.92$\pm$0.06}& \textbf{0.91$\pm$0.07} & \textbf{0.90$\pm$0.08} &\textbf{0.94$\pm$0.09} & 0.88$\pm$0.10  \\ 
        \hline
        \end{tabular}
        }
    \end{center}
\end{table}

\noindent\textbf{Results.} 
All performances averaged over 5 repetitions are reported in Table~\ref{tab2}. They underline the clear benefits of HyperMM for cancer classification from histopathological images. In particular, our method outperforms magnification-specific models, and is closely followed by Incremental-CNN, which highlights the benefits of combining the information carried by different magnifications. Moreover, while Incremental-CNN maximises the recall score of the task, HyperMM maximises precision, and overall improves upon Incremental-CNN. This shows that learning to predict an early latent combination of features (i.e. combining multiple images of a same patient during model training directly) yields better performances than combining predictions made on individual images. \\

\noindent\textbf{Discussion.}
While the analysis of multi-resolution images is not a multimodal application by definition, our method is designed to enable mid-level fusion of latent features of varying-sized inputs, and is therefore adapted for this use case. Because of the varying number of images per patient in histopathology datasets, traditional approaches are not equiped to combine multiple resolutions directly during training to make patient-level decisions, and instead rely on the late fusion of image-level decisions. In contrast, HyperMM offers this possibility. It opens a new and different way to classify cancer patients. Moreso, our experiments suggest that mid-level fusion even considerably improves the performances of existing late fusion models.

\section{Conclusion}
We have demonstrated the advantages of HyperMM for robust MML with missing modalities: our method eliminates the need to use complex and computationally costly imputation strategies, thus significantly decreasing model training time; and unlike competitors, its performances are not dependant on which modality is missing in the data. In particular, unlike imputation-based methods, our approach is end-to-end: HyperMM eliminates the time-consuming steps of manually imputing the missing modalities using a previously trained imputation model, before finally training a prediction model. On the contrary, our two-step model is trained without interruption or human intervention. Moreover, by only utilising the observed images of the incomplete dataset, HyperMM avoids prediction bias caused by poor imputation or the presence or generated dummy data. In addition, we have shown that the flexibility of HyperMM alleviates the constraints usually met in applications with varying-sized datasets and opens up a whole new range of possible learning strategies. Our framework is \textit{task-agnostic}, and can be easily used beyond the two applications we have presented. For instance, it could be extended to multivariate time series analysis, where incomplete data is common (e.g. damaged channels in EEG recordings). Moreover, while we used pre-trained feature extractors in all our experiments for simplicity, HyperMM is \textit{model-agnostic} and adaptable to any  neural network-based feature extractor or predictor.

\subsection*{Acknowledgements} 
This work has been supported by the French government, through the 3IA Côte d’Azur Investments in the Future project managed by the National Research Agency (ANR) (ANR-19- P3IA-0002).

%
%
%
%

\bibliography{Main/main}

\end{document}